\documentclass[a4paper,twoside]{article}      
\usepackage{amsmath,amssymb,amsfonts} 
\usepackage{graphics}                 
\usepackage{color}                    
\usepackage{hyperref}                 
\usepackage{comment} 			
\usepackage{enumitem}

\date{}
\title{What Really is Deep Learning Doing?
\footnote{Great thanks for whole heart support of my wife.}}

\author{ Chuyu Xiong \\
{\small Independent Researcher, New York, USA} \\
{\small Email: chuyux99@gmail.com}
}

\begin{document}
\maketitle
\begin{abstract}
Deep learning has achieved a great success in many areas, from computer vision to natural language processing, to game playing, and much more. Yet, what deep learning is really doing is still an open question. There are a lot of works in this direction. For example, \cite{renormal} tried to explain deep learning by group renormalization, and \cite{why} tried to explain deep learning from the view of functional approximation. In order to address this very crucial question, here we see deep learning from perspective of mechanical learning and learning machine (see \cite{paper1},  \cite{paper2}). From this particular angle, we can see deep learning much better and answer with confidence: What deep learning is really doing? why it works well, how it works, and how much data is necessary for learning. We also will discuss advantages and disadvantages of deep learning at the end of this work.   
\end{abstract}


\section{Introduction}\label{table}
In recent years, deep learning (a branch of machine learning) has achieved many successes in lot of fields. However, a clear theoretical framework of deep learning is still missing. Consequently, there are many fundamental questions about deep learning are still open. For example: What is deep learning really doing? Is it really learning, or just a kind of fancy approximation to a function? Why it indeed has so many success? Why deep learning needs big data? For a particular problem, how much data is sufficient to drive deep learning to learn? Up to now, there is no satisfactory answer for these fundatamental questions. We here are trying to address these questions from a new angle.  

We introduced term "Mechanical learning" in \cite{paper1}. Mechanical learning is a computing system that is based on a simple set of fixed rules (this is so called mechanical), and can modify itself according to incoming data (this is so called learning). A learning machine is a system that realizes mechanical learning. 

In \cite{paper2}, we described learning machine in a lot of details. By doing so, we gained some useful knowledges and insights of mechanical learning. 

In this short article, by using those knowledges and insights, we are trying to view deep learning from a new angle. First, we will briefly discuss learning machine, pattern, internal representation space, X-form, data sufficiency, and learning strategies and methods. Then, we will use the view of learning machine to see deep learning. We start from the simplest deep learning, i.e., 2-1 RBM, then go to 3-1 RBM, $N$-1 RBM, $N$-$M$ RBM, stacks of RBMs. Then we discuss the learning dynamics of deep learning. By this approach, we see clearly what deep learning is doing, why deep learning is working, under what condition deep learning can learn well,  how much data are needed, and what disadvantages deep learning has.

\section{Mechanical Learning and Learning Machine}\label{table}
We here very briefly sum up the discussions that we did in \cite{paper1} and \cite{paper2}. 
A learning machine $\mathcal{M}$ has 2 major aspects: it is an IPU, i.e. it is able to process information; and it is learning, i.e. its information processing ability is changing according to data. Without learning (since it is a machine we design, we can stop learning), $\mathcal{M}$ is very similar to a CPU. However, one major difference between learning machine and CPU is: learning machine treat incoming data according to its pattern, not bit-wise.  

Thus, in order to understand a learning machine, it's absolutely necessary to understand pattern well. There are 2 kinds of patterns: objective pattern and subjective pattern. Subjective pattern is crucial for learning machine. In \cite{paper2}, we proved one theorem: For any objective pattern $P_o$, we can find a proper subjective pattern $p$ that can express $P_o$ well and is build upon a least set of base patterns. To describe subjective patterns, it is best to use X-form, which is one algebraic expression upon some base patterns. X-form is one very important mathematical object. X-form could have sub-forms. X-form and its sub-forms actually forms the fundamental fabric of a learning machine.

We also defined learning by teaching and learning without teaching. Then, further specify {\it typical mechanical learning}. Learning by teaching requires we know learning machine and the pattern to learn well. By these knowledges, we can design a teaching sequence to make learning machine learn well. We proved that if a learning machine has certain capabilities for learning by teaching, it is universal, i.e. able to learn anything. 

However, most learning is not learning by teaching. In order to understand typical mechanical learning, we introduced {\it internal representation space}. Structurally, a learning machine $\mathcal{M}$ has components: input space, output space, internal representation space, and learning methods and learning strategies. The most important part is internal representation space. We studies internal representation space $\mathcal{E}$ in details, and revealed, in fact, it is equivalent to a collection of X-forms. This fact tells us that learning is nothing but a dynamics on $\mathcal{E}$, moving from one X-form to another. With clear and reachable internal representation space, learning can be understood much better, and can be done much more efficiently. For example, we can unify all 5 kinds of learning -- logic reasoning, connectionism, probabilistic approach, analogy, and evolution (see \cite{pedro}) --  together on $\mathcal{E}$ naturally.

For mechanical learning, we need to understand data sufficiency. This is very crucial concept. We use X-form and its sub-forms to define data sufficient to support one X-form and sufficient to bound one X-form. With sufficient data, we can see how learning strategy and learning method work. There could be many learning strategies and learning methods. We show 3 learning strategies: 1. Embed X-form into parameter space. 2. Squeeze X-form from inside to higher abstraction. 3. Squeeze X-form from inside and outside to higher abstraction. We prove that with certain capabilities, the last 2 strategies and methods will make universal learning machine.  Of course, this is theoretical results, since we have not designed a specific learning machine yet. 

Here, we will show that deep learning is actually doing mechanical learning by the first strategy, i.e. embed X-forms into parameter space. Such a fact will help us to understand deep learning much better.

\section{See Deep Learning from View of Learning Machine}\label{table}
According to our definition, if without human intervention, deep learning is mechanical learning. Of course, this "if" is a big if. Often, deep learning program is running with a lot of human intervention, specially at the time of model set up. We will restrict our discussion to Hinton's original model \cite{hinton}, i.e., a stack of RBMs. Each level of RBM is clearly a $N$-$M$ learning machine ($N$ and$M$ are dimensions of input and output). Hinton's deep learning model is by stacking RBM together. If without further human intervention, it is a learning machine. This is the original model of deep learning. Other deep learning program can be thought as variations based on this model. Though in the past few years deep learning has leaped forward greatly, stacking RBM together still reflects the most typical properties of deep learning.  

Thus, we would expect many things in mechanical learning could be applied to deep learning. The point is, we are viewing deep learning from a quite different angle -- angle of mechanical learning. For example, we can view Hinton's original deep learning program \cite{hinton} as one 258-4 learning machine, and we ask what is the internal representation space of it? We expect such angle and questions would reveal useful things for us.
The central questions indeed are: what is the internal representation space of deep learning? what is the learning dynamics? At first, seems it is quite hard since learning is conducted on a huge parameter space (dimension could be hundreds of millions), and learning methods are overwhelmingly a big body of math. However, when we apply the basic thoughts of learning machine to deep learning, starting from simplest RBM, i.e. 2-1 RBM, we start to see much more clearly.
\bigskip

{\bf 2-1 RBM} \\
2-1 RBM is the simplest. However, it is also very useful since we can examine all details, and such details will give us a good guide on more general RBM. 

2-1 RBM is one IPU. We know 2-1 IPU totally has 16 processing ($2^{2^2}$). But, we only consider those processing: $p(0, 0) = 0$, so totally 8 processing, which we denote as $P_j, j=0, \ldots, 7$ (see \cite{paper1}). For 2-1 RBM, any processing $p$ can be written as: for input $(i_1, i_2)$, the output $o$ is:
$$
o = p(i_1, i_2) = \text{Sign}(a i_1 + b  i_2), \text{where} \ (a, b) \in \mathbb{R}^2, \ \text{Sign}(x) =
\left\{
	\begin{array}{ll}
		1  & \mbox{if } x \ge 0 \\
		0  & \mbox{if } x < 0
	\end{array}
\right.
$$


The parameters $(a, b)$ determine what the processing really is. Parameter space $\mathbb{R}^2$ has infinite many choices of parameters. But, there are only 6 processing, thus, for many different parameters, the processing is actually same. We can see all processing in below table:

\begin{center}
\centering
    \begin{tabular}{|c|c|c|c|c|c|c|c|c|c|c|c|c|c|c|c|c|}
        \hline
        ~     & $P_0$ & $P_1$ & $P_2$ & $P_3$ & $P_4$ & $P_5$ & $P_6$ & $P_7$  \\ \hline
        (0,0) & 0     & 0     & 0     & 0     & 0     & 0     & 0     & 0       \\ 
        (1,0) & 0     & 1     & 0     & 1     & 0     & 1     & 0     & 1       \\ 
        (0,1) & 0     & 0     & 1     & 0     & 1     & 1     & 0     & 1       \\ 
        (1,1) & 0     & 0     & 0     & 1     & 1     & 0     & 1     & 1       \\
        \hline
          Region   & $R_4$ & $R_3$ & $R_5$ & $R_2$ & $R_6$ & None & None & $R_1$   \\	
        \hline
          X-form    & $0$ & $b_1$ & $b_2$ & $b_1 + b_3$ & $b_2 + b_3$ & $b_1 + b_2$ & $b_3$ & $b_1 + b_2 + b_3$   \\	
        \hline
    \end{tabular}

{\bf Tab. 1.  Table for all processing of 2-1 RBM} 
\end{center}


\begin{center}
\begin{picture}(220,250)(0,0)
\put(0, 10){\resizebox{9 cm}{!}{\includegraphics{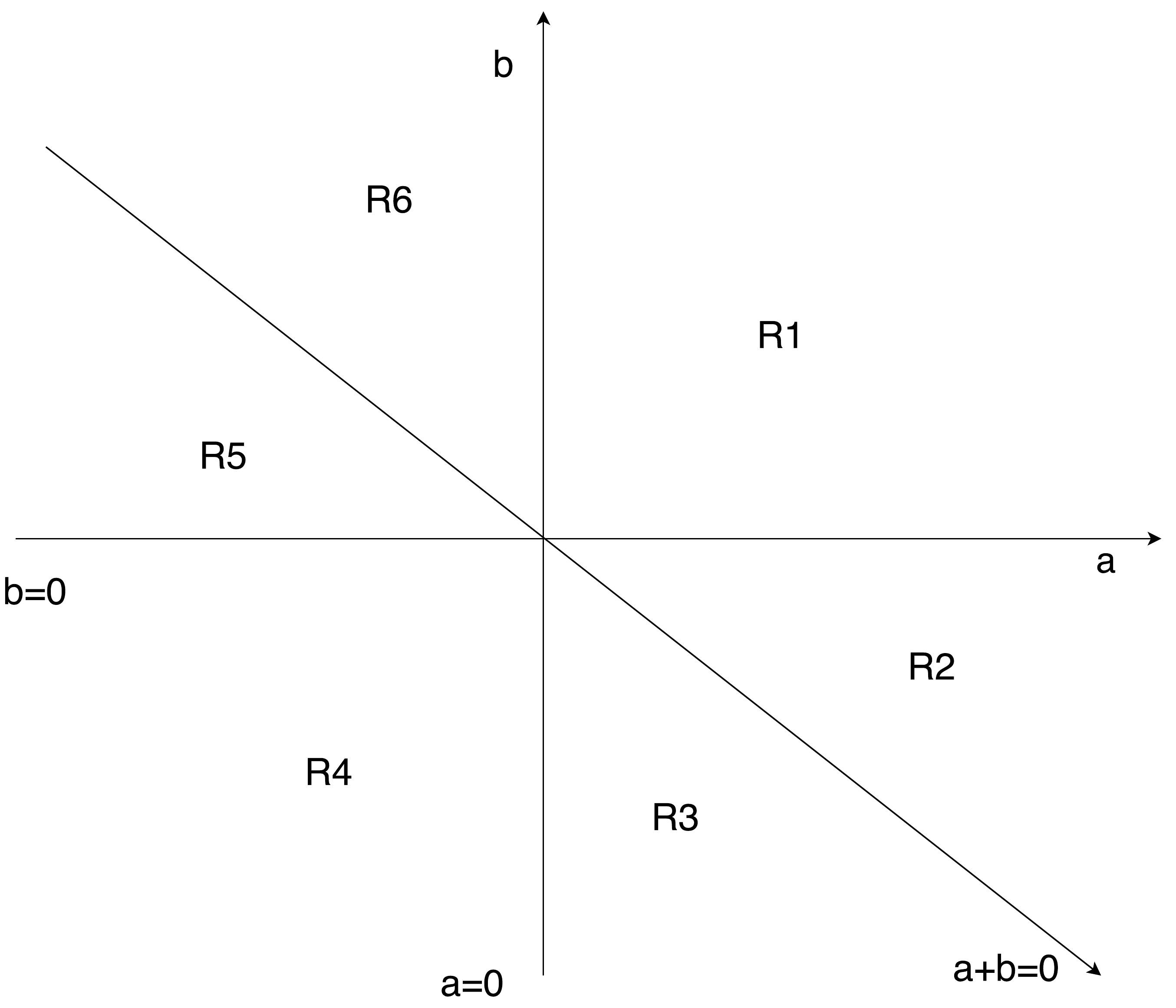}}}
\end{picture}

{\bf Fig. 1. Parameter space of 2-1 RBM that is cut into 6 regions} 
\end{center}

In first row of table, $P_i, i = 0, \ldots, 7$ are all processing of 2-1 IPU. Under first row, there is value table for each processing. We point out some quite familiar processing: $P_7$ is actually logical OR gate, $P_6$ is logical AND gate, $P_5$ is logical XOR gate. Note, $P_5, P_6$ are processing for 2-1 IPU, but not in 2-1 RBM. It is well known, 2-1 RBM has no XOR and AND (i.e. no $P_5, P_6$).

$R_j$, $j=1, \ldots, 6$ indicate regions in parameter space $\mathbb{R}^2$, each region for one processing. There are only 6 regions, since 2-1 RBM only has 6 processing. We briefly discuss how we get these regions. See illustration in figure.

Suppose $p$ is processing. Line $a = 0$ cuts $\mathbb{R}^2$ into 2 regions: $a \ge 0$ and $a < 0$. If $(a, b)$ is in first region, then $p(1, 0) = 1$, in second, $p(1, 0) = 0$. Line $b = 0$ is perpendicular to $a=0$, so, it cuts the previous 2 regions into 4 regions: $a \ge 0, b \ge 0$ and $a \ge 0, b < 0$ and  $a < 0, b \ge 0$ and $a < 0, b < 0$. Clearly, if $b \ge 0$, $p(0, 1) = 1$, if $b < 0$, $p(0, 1) = 0$. Line $a + b = 0$ could no longer cuts the previous 4 regions into 8 regions, it could only cut 2 regions (2nd, 4th quadrant) into 4 regions ($R_2, R_3, R_5, R_6$). So, totally, we have 6 regions, and each region is for one processing. This argument about regions is very simple, yet very effective. We can extend this argument to $N$-1 RBM.
 
Each region is for one processing. So, region can be used to represent processing. That is 6th row in the table shown. Yet, a much better expression is by X-form (\cite{paper2}). We explain them here. 
Here $b_0 = (0, 0), b_1 = (1, 0), b_2 = (0, 1), b_3 = (1, 1)$ are base patterns. For 2-dim pattern space, there are only these 4 base patterns. But, $b_i$ can also be used to represent one processing of 2-1 IPU, i.e. $b_i$ is such a processing: when input is $b_i$, output is 1, otherwise output is zero. X-forms are expressions based on all $N$-1 base patterns, operations +, $\cdot$, $\neg$, composition, Cartesian product, and apply them consecutively. Example, $b_1 + b_3$, $b_1 \cdot b_2$, $b_1 + \neg b_2$ are X-forms. 
Any processing of 2-1 IPU can be expressed by at least one X-form \cite{paper2}. For example, if region is $R_3$, processing is $P_1$, X-form is $b_1$. 
Another example, region is $R_1$, processing is $P_7$ (this is OR gate), X-form is $b_1 + b_2 + b_3$. $P_5$ is a processing of 2-1 IPU (XOR gate), but not in 2-1 RBM, its X-form is $b_1 + b_2$.  
The 7th row in the table shows X-forms representing processing. We can say that each processing is represented by a region, and by a X-form as well.

When 2-1 RBM is learning, clearly, parameter $(a, b)$ is adapting. But, only when $(a, b)$ cross region, processing changes. Before crossing, change of parameters is just for preparation for crossing (perhaps many parameter changes are just wasted). Learning is moving from one region to another. Or, equivalently, learning is from one X-form to another. Such view is crucial. Now, we are clear, on surface, learning on 2-1 RBM is a dynamics on parameter space $\mathbb{R}^2$, but real learning dynamics is on 6 regions (or X-forms). Such indirectness causes a lot of problem. 
\bigskip

{\bf 3-1 RBM} \\
Just increase input dimension 1, we have 3-1 RBM. To discuss it, we can gain some insights for general RBM. For 3-1 RBM, still we can write: for any input $(i_1, i_3, i_3) \in \mathbb{B}^3$, output $o \in \mathbb{B}$ is:
$$
o = p(i_1, i_2, i_3) = \text{Sign}(a i_1 + b  i_2 + c i_3), \text{where} \ (a, b, c) \in \mathbb{R}^3, \ \text{Sign}(x) =
\left\{
	\begin{array}{ll}
		1  & \mbox{if } x \ge 0 \\
		0  & \mbox{if } x < 0
	\end{array}
\right.
$$

However, while we can easily write down all possible processing of 2-1 RBM, it would be hard to do so for 3-1 RBM. For 3-1 IPU, we know that the number of all possible processing is $2^{2^3} = 2^8 = 256$. Since only considering such processing $p$: $p(0, 0, 0) = 0$, the number becomes $256/2 = 128$. We expect 3-1 RBM has less processing. But, how many possible processing of 3-1 RBM could have?

Following the guidance that 2-1 RBM gives to us, i.e. to consider the hyperplane generated by nonlinearity that cuts parameter spaces, we examine parameter space $(a, b, c) \in \mathbb{R}^3$, and following planes:
$a = 0, b = 0, c = 0, a+b = 0, a+c = 0, b+c = 0, a+b+c =0$. 
These planes are naturally obtained. For example, if we consider the input $(1, 0, 0)$, it is easy to see plane $a = 0$ is where cut the value of output: 1 or 0. So, the above planes are associated with following inputs:
$(1, 0, 0), (0, 1, 0), (0, 0, 1), (1, 1, 0), (1, 0, 1), (0, 1, 1), (1, 1, 1)$

We can clearly see that in one region that is cut out by above 7 planes, the output values are same. Therefore, one region actually is representing one processing: in the region, processing is same. So, question of how many possible processing becomes how many possible regions cut out by the 7 planes. We do counting for the regions below. 

First, $a = 0$ cuts parameter space into 2 pieces: $R_1^1$, $R_2^1$. 
Second, $b = 0$ perpendicular to $a = 0$, so, it cuts each region  $R_1^1$, $R_2^1$ into 2 pieces, we then have 4 regions: $R_1^2$, $R_2^2$.$R_3^2$, $R_4^2$. 
Then, $c = 0$ perpendicular to $a = 0$ and $b = 0$, so, we have 8 regions: $R_j^3, j=1, \ldots, 8$.
Then, consider $a+b = 0$. This plane no longer could be perpendicular to all $a = 0, b = 0, c = 0$. We will not have $2 * 8 = 16$ regions. We only have $1.5 * 8 = 12$ regions.
Following the same argument, we have: For $a+c = 0$, $1.5 * 12 = 18$ regions. For $b+c = 0$, $1.5 * 18 = 27$ regions. For $a+b+c = 0$, $1.5 * 27 < 41$ regions. 

So, for 3-1 RBM, there are at most 41 possible processing, comparing to 128 possible processing of full 3-1 IPU. However, there are possibility that the number of processing is even less than 41, since among those regions, it is possible that 2 different regions give same processing. We do not consider these details here. 

Since regions can be represented by X-form, each processing 3-1 RBM can be represented by at least one X-form.  $b_1 = (1,0,0), b_2 = (0,1,1), b_3=(0,0,1), \ldots, b_7 = (1,1,1)$ are X-form for all base patterns. Examples, X-form $b_1 + b_2$ is in 3-1 RBM.  But, $b_1 \cdot b_2$ is not in 3-1 RBM. There are a lot of such X-form that is not in 3-1 RBM.

Learning dynamics on 3-1 RBM is also in such way: on surface, it is dynamics on $\mathbb{R}^3$, but real learning dynamics is on 41 regions (or X-forms). 
\bigskip

{\bf $N$-1 RBM} \\
The argument for 3-1 RBM can be extended to $N$-1 RBM (See details in \cite{paper2}). We consider hyperplanes and regions cut off by these hyperplanes. The number of these regions is less than:  $2^N 1.5^{2^N-N-1}$. Compare to the number of all processing of $N$-1 IPU, which is $2^{2^N - 1}$, easy to see, $N$-1 RBM has much less processing. This means that $N$-1 RBM could not express many processing. 

For $N$-1 RBM, still we can write: for any input $(i_1, \ldots, i_N) \in \mathbb{B}^N$, output $o \in \mathbb{B}$ is:
$$
o = p(i_1, \ldots, i_3) = \text{Sign}(a_1 i_1 + a_2  i_2 + \ldots + a_N i_3), \text{where} \ (a_1, \ldots, a_N) \in \mathbb{R}^N, \ \text{Sign}(x) =
\left\{
	\begin{array}{ll}
		1  & \mbox{if } x \ge 0 \\
		0  & \mbox{if } x < 0
	\end{array}
\right.
$$

\begin{equation}
a_1 = 0, \ldots, a_1+a_2 = 0, \ldots, \ a_1+a_2+a_3 =0 \ldots, \ \ldots, \ a_1+a_2+a_3+\ldots = 0, \ldots
\end{equation}
There are $ {N\choose 1} $ hyperplanes such as $a_1 = 0$;  $ {N\choose 2} $ hyperplanes such as $a_1+a_2 = 0$;
\ldots. We also have this: First $N$ hyperplanes will cut parameter space into $2^N$ regions. Then, later each hyperplanes will cut more regions by the rate of multiplying factor $1.5$. Thus, we can see the number of regions are:

$$
2^N * 1.5^{K_2} * 1.5 ^{K_3} * \ldots * 1.5^{K_N}
$$
where ${K_2} = {N\choose 2}$ is the number of hyperplanes such as $a+b = 0$, etc. 

And, we have the equation:

\begin{equation}
2^N = {N\choose 0} +  {N\choose 1} +  {N\choose 2} + \ldots +  {N\choose N}   
\end{equation}

So, 

\begin{equation}
K_2 + K_3 + \ldots + K_N = {N\choose 2} +  {N\choose 3} + \ldots +  {N\choose N} = 2^N - {N\choose 1} -{N\choose 0}    = 2^N - N -1
\end{equation}

Thus, the number of regions are

$$
2^N * 1.5^{2^N-N-1} = 2^N * ({\frac{3}{2}})^{2^N-N-1} 
$$

This is a very big number. Yet, compare to the total possible processing of full IPU, it is quite small. See their quotient:

\[
\frac{2^{2^N}}{2^N * ({\frac{3}{2}})^{2^N-N-1} }   = 2* (4/3)^{2^N - N - 1}
\]
It tells that full IPU has $f = 2* (4/3)^{2^N - N - 1}$ times more processing comparing to RBMs. This is huge difference. Say, just for $N = 10$, $f$ is more than 120 digits, i.e. the number of processing of full IPU would has more 120 digits at the end than the number of RBMs.

Also, each region can be expressed by at least one X-form. For examples, $b_1 + b_N$, $b_1 + b_3 + b_5$, etc. Learning dynamics on $N$-1 RBM is in such way: on surface, it is dynamics on $\mathbb{R}^N$, but real learning dynamics is on those  $2^N 1.5^{2^N-N-1}$ regions (or X-forms). 
\bigskip

{\bf $N$-$M$ RBM} \\
Suppose $\mathcal{R}_i, i=1, \ldots, M$ are $M$ $N$-1 RBMs, we can form a $N$-$M$ RBM, denote as $\mathcal{R} = (\mathcal{R}_1, \ldots. \mathcal{R}_M)$, whose processing are  $p = (p_1, p_2, \ldots, p_M)$, where $p_i, i = 1, \ldots, M$ are processing of $\mathcal{R}_i$. So, $\mathcal{R}$ is Cartesian product of $\mathcal{R}_i, i=1, \ldots, M$. 

Since all $\mathcal{R}_i$ are cut into regions, and in each region, processing is same, we can see $\mathcal{R}$ is also cut into regions, and each region is a Cartesian product of regions of $\mathcal{R}_i$: $R = R_1 \times R_2 \times \ldots \times R_M$, where $R_i$ is one region from i-th RBM $\mathcal{R}_i$. Thus, the number of all possible regions of $\mathcal{R}$ is $(2^N 1.5^{2^N-N-1})^M = 2^{NM} 1.5^{M(2^N-N-1)}$. This is a much smaller number than $2^{M2^N}$, which is the number of total possible processing for $N$-$M$ IPU. 

X-form for each region of $\mathcal{R}$, are actually Cartesian product of X-form for those regions of $\mathcal{R}_i$. Suppose $R = R_1 \times R_2 \times \ldots \times R_M$, and $f_i$ are X-forms for region $R_i$ in $\mathcal{R}_i$, $i=1, \ldots, M$, then X-form for $R$ is $f = (f_1, \ldots, f_M)$.  For example, $(b_1, b_1+b_3, \ldots, b_2 \cdot b_4)$ is one X-form of $\mathcal{R}$. 

Learning dynamics on $N$-$M$ RBM is in such way: on surface, it is dynamics on parameter space $\mathbb{R}^{NM}$, but real learning dynamics is on those  $2^{NM} 1.5^{M(2^N-N-1)}$ regions (or X-forms). 
\bigskip

{\bf Stacking RBMs} \\
Consider a $N$-$M$ RBM $\mathcal{R}_1$, and a $M$-$L$ RBM $\mathcal{R}_2$, stacking them together, we get one $N$-$L$ IPU $\mathcal{R}$: A processing $p$ of $\mathcal{R}$ are composition of processing $p_1, p_2$ of $\mathcal{R}_1, \mathcal{R}_2$: $p(i_1, i_2, \ldots, i_N) = p_2 ( p_1(i_1, i_2, \ldots, i_N) )$. And we denote as: $\mathcal{R} = \mathcal{R}_1 \otimes \mathcal{R}$.

The parameter space of $\mathcal{R}$ clearly is $\mathbb{R}^{NM} \times \mathbb{R}^{ML}$. We know $\mathbb{R}^{NM}$ is cut into some regions, in each region processing is same. So does $\mathbb{R}^{ML}$. Thus, $\mathbb{R}^{NM + ML}$ is cut into some regions, in each region processing is same, and these regions are Cartesian product of regions in $\mathbb{R}^{NM}$ and $\mathbb{R}^{ML}$. So, we know number of total possible processing $\mathcal{R}$ equals total possible processing of $\mathcal{R}_1$ times $\mathcal{R}_2$, i.e. 
$2^{NM} 1.5^{M(2^N-N-1)} \times 2^{ML} 1.5^{L(2^M-M-1)} = 2^{NM+ML} 1.5^{M(2^N-N-1) + L(2^M-M-1)}$. 

We can easily see if $M$ is large enough, the above number will become greater than $2^{L2^N}$, which is total possible processing of $\mathcal{R}$. We can see, at least potentially, $\mathcal{R}$ has enough ability to become a $N$-$L$ full IPU. But, we will not consider here. In fact, it is very closely related to the so called Universal Approximation Theorem. Indeed, stacking RBM together is powerful. 

X-form can be expressed as composition as well. For example, consider 3 2-1 RBM $\mathcal{R}_1$, $\mathcal{R}_2$, and $\mathcal{R}_3$. Using $\mathcal{R}_1$ and $\mathcal{R}_2$ to form a 2-2 RBM, and using $\mathcal{R}_3$ to stack on it, we get a 2-1 IPU $\mathcal{R}$: $\mathcal{R} = \mathcal{R}_3 \otimes (\mathcal{R}_1, \mathcal{R}_2)$. If for this case, $\mathcal{R}_1$ has X-form $b_1$, and $\mathcal{R}_2$ has X-form $b_2$, and $\mathcal{R}_3$ has X-form $b_1 + b_2 + b_3$, them, $\mathcal{R}$ has X-form $(b_1 + b_2 + b_3) (b_1, b_2)$. Easy to see this X-form is processing $P_5$ (XOR gate), which is not expressible by one 2-1 RBM. So, putting 3 2-1 RBMs together, more X-form can be expressed.

\section{Learning Dynamics of Deep Learning}\label{table}
With these understandings RBMs, which is the most essential building block of deep learning, we can see how the model of deep learning is build up, and how learning dynamics is doing. Clearly, today's deep learning is much more than original Hinton's model of stacking RBMs (see \cite{hinton}). But, we would first talking about this model.

The deep learning model is by stacking more RBMs (this is so called deep). Once several RBMs are putting together, a deep learning model is formed. Suppose $\mathcal{R}_j$, are $N_j$-$N_{j+1}$ RBM, $j = 1, \ldots, J$, where $N_1, \ldots, N_J, N_{J+1}$ are sequence of integers.  So, we can stack these RBM together to form one $N_1$-$N_{J+1}$ IPU, whose processing could be written in such way: $p(i_1, i_2, \ldots, i_{N_1}) = p_J ( \ldots  p_2 ( p_1(i_1, i_2, \ldots, i_{N_1}) )$
where each $p_j$ is processing of $\mathcal{R}_j$. We denote this IPU as $\mathcal{R} =  \mathcal{R}_1 \otimes \ldots \otimes \mathcal{R}_J$.  All parameters of  $\mathcal{R}$ form a huge Euclidean space $\mathbb{R}^{N_1N_2 + \ldots + N_JN_{J+1}}$. We can denote this huge parameter space as $\mathbb{R}^*$.

Then, clearly, deep learning is conducted on $\mathcal{R}$ to reach a good processing by modifying the parameters in $\mathbb{R}^*$. Of course, it requires skills to modify such a huge number of parameters. There are methods, such as CD (convergent divergence), SGD (stochastic gradient descent), etc. are invented for such purpose. 

However, no matter what methods are used to modify the parameters, it is modifying parameters to form the dynamics of learning. So, seems the phase space of learning dynamics is on the space $\mathbb{R}^*$. But, this is just on surface. As we discussed in last section, the true dynamics is not conducted on parameters, but on those regions. The learning dynamics is conducted on these regions cut by hyperplanes and Cartesian products.  The number of regions are huge: $2^{N_1N_2} 1.5^{N_2(2^{N_1}-N_1-1)} \times \ldots$. 


More precisely, the situation is: as learning, a huge number of parameters are adapting, but only when parameters cross region, the processing of $\mathcal{R}$ changes. Before crossing, processing remains same, the changes of parameter at most can be thought as the preparation for crossing (perhaps many such  changes of parameters are just wasted). Thus, learning dynamics is to move from one region to another. We also know each region is associated with one X-form. Thus, learning dynamics is to move from one X-form to another X-form.

\begin{center}
\begin{picture}(220,220)(0,0)
\put(0, 10){\resizebox{7 cm}{!}{\includegraphics{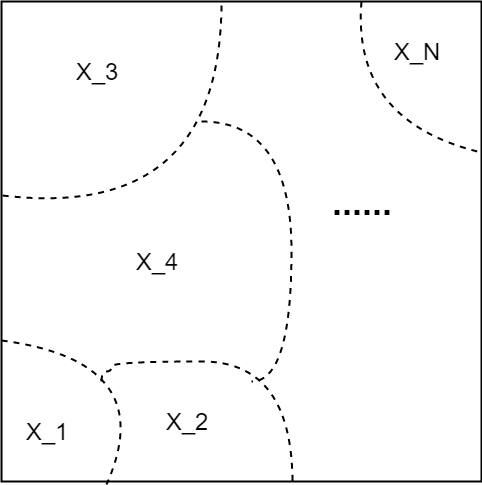}}}
\end{picture}

{\bf Fig. 2. Illustration on parameter space is cut into regions} 
\end{center}

Fig. 2 gives one illustration on parameter space is cut into regions. Of course,  $\mathbb{R}^*$ is very high dimension Euclidean space, so regions could be shown on paper precisely, and Fig. 2 is just one illustration. However, this illustration gives us one clear picture about deep learning structure.  

The deep learning structure is formed by these factors: how many RBMs, dimension of each RBM, how to stack, how to do Cartesian product. Once the structure is formed, if no further human intervention (such as manually adjust numbers or subroutines in model), the structure will not change. Such structure is formed by people at set up time. So, for fixed structure, we will have fixed region cut (as illustrated in Fig. 2). Further, we will have a fixed set of X-forms, and learning is conducted on this set of X-forms.


We can see one example. $\mathcal{R}_1$, $\mathcal{R}_2$, $\mathcal{R}_3$ are 3 2-1 RBMs. We put them like this: $\mathcal{R} = \mathcal{R}_3 \otimes (\mathcal{R}_1, \mathcal{R}_2)$. We have 3 parameter space $(a, b), (c, d), (e, f)$. We have 6 regions for each parameter space. Put them together, we have 6x6x6 = 216 regions. $\mathcal{R}$ is one 2-1 IPU. So, $\mathcal{R}$ at most has 8 processing. Thus, among those 216 regions some different regions will have same processing. But, each region will have one X-form. That is to say, for one processing, there could have several X-form associated with. For example, consider this region: $R^3_1 \times (R^1_3, R^2_5)$. This gives processing $P_5$ (XOR gate). Normally, for this processing, we can use X-form $b_1 + b_2$ for it. But, for the region, naturally, the X-form is: $(b_1 + b_2 + b_3) \otimes (b_1, b_2)$. That is to say, this X-form will generate the same processing as $b_1 + b_2$. Another region: $R^3_2 \times (R^1_2, R^2_6)$ will give the same processing. And, the X-form is: $(b_1 + b_3) \otimes (b_1+b_3, b_2+b_3)$.

For deep learning model build on stacking RBMs (original Hinton's model, \cite{hinton}), as we discussed in last section, the situation is same: the parameter space $\mathbb{R}^*$ is cut into regions (by hyperplanes, etc), and each region is associated with one X-form, when parameter cross the boundary of the regions, one X-form moves to another X-form, learning dynamics is conducted on this set of X-form. 


This is the learning dynamics of deep learning, this is what really deep learning is doing. 

For more complicated deep learning, such as convolution is used, connection pruning is done, nonlinearity is other than sign function (like ReLU), the situation will be more complicated. However, if there is no human intervention, it will surely be mechanical learning. We still can prove that the learning dynamics is the same:  the parameter space $\mathbb{R}^*$ is cut into regions, and each region is associated with one X-form, when parameter cross the boundary of the regions, one X-form moves to another X-form, learning dynamics is conducted on this set of X-form. 

To prove this for general deep learning mathematically, additional works are needed. We will do this work in other place. But, we have no doubt this can be done. 

Such a learning strategy is exactly what we described in \cite{paper2}: "Embedded X-forms into parameter space".

\section{Remark}\label{table}
Now, we know the fact: deep learning is using strategy of "Embedded X-forms into parameter space". This fact is very essential and many consequences can be derived from it. Here we make some comments.

{\bf True nature of deep learning: }\\
On surface, deep learning seems build a model from data feed into eventually (by using neural network, stacking RBMs, and more other tools). However, as we reveal in previous sections, it is not such a case. Essentially, a deep learning model is doing this: at the time of model setting up, to cut the huge parameter space  $\mathbb{R}^*$ into many regions, and each region is associated with one X-form that is one logical statement, then driven by big amount of data, following certain learning dynamics, i.e. to move from region to another region, which is equivalent to move from one X-form to another X-form, and eventually to reach a satisfactory X-form, which is the learning result.

So, we say that deep learning is not to building up a model from data input, but it is to choose a good X-form from a set of X-forms established at the time deep learning model is set up. This is the true nature of deep learning.

Such a view is different than popular view about deep learning. However, it is true and help us to understand deep learning better. For example, \cite{renormal} might be right, there are some group renormalization going on, but, it missed this issue: group renormalization happens at the setting stage not at the learning stage. Another example, \cite{why} gives a very good explanation about the power of multi-stage-composition. However, it failed to realize that learning is not only to get a good processing, but to find a best possible X-form for the good processing (since one particular processing could have many associated X-form, and some of such X-form is bad, some of such X-form is good).

 {\bf Fundamental limitation of deep learning: }\\
The fundamental limitation of a deep learning model is from its nature: it acts on a pre-selected set of X-forms $\mathfrak{X}$, which is formed at the time the model is set up. 

So, mostly likely, a deep learning model could not be an universal learning machine \cite{paper2}. If it is universal, $\mathfrak{X}$ must contains at least one X-forms for any possible processing. This is nearly impossible. 


Actually, a deep learning model is set up by human, and is for one particular task. So, most likely, $\mathfrak{X}$ only contains X-forms specially for this task. And, this deep learning model is limited by $\mathfrak{X}$.

If the learning target is for a particular processing, and if in $\mathfrak{X}$ there is at least one X-form associated to this processing, a deep learning model could possible to get the target. Otherwise, a deep learning model could not reach the processing, no matter how hard to try and how much data. In another word, the model is a bad model. But, deep learning has no any method to tell if a model is good or bad before trying it out. This is one huge limitation.

Yet, even $\mathfrak{X}$ contains a X-form associated with the desired processing, we still do not know if the X-form is a good one. As we know in \cite{paper2}, there are many X-forms associated with one processing, some is bad, and some is good. For example, one X-form is more robust for certain conditions. If $\mathfrak{X}$ does not contain the robust X-form, no matter how hard to try and how much data, learning could not get a robust solution. Again,  deep learning has no any method to tell if $\mathfrak{X}$ contains such X-form. . 

These limitations are fundamental and are derived from the fact: deep learning is chosen X-form from a pre-selected set  $\mathfrak{X}$, not dynamically building a X-form from input data.

{\bf Logical or Probabilistic: }\\
Quite often, people think deep learning is doing probability learning. They think that a probability model is essential for deep learning since a lot data feed in, specially, stochastic gradient descent is one very essential learning method. However, we would like to point out: deep learning fundamentally is viewing its learning target logically. Why? Each X-form in $\mathfrak{X}$ is a logical statement, often a very long logical statement (so called deep). So, the very essential thing is: when a deep learning model does its information processing, it is doing according to a solid logical statement. Not doing the information processing probabilistically.

Of course, the way to get the X-form might not be pure logical, it could involve a lot of probabilistic views and actions, such as stochastic gradient descent. However, we would point out, even the way to get the X-form, could be pure deterministic not probabilistic. It is possible to design a pure deterministic learning dynamics, at least in theory.

 {\bf Why works well: }\\
Practice shows, deep learning works very well for many problems. Now, we can see the reason for such success clearly: when a deep learning model for one problem is set up by experienced people, the desired X-form is already build into the model. More precisely, if the desired X-form is $X$, when deep learning model is set up,  we have $X \in \mathfrak{X}$, where $\mathfrak{X}$ is the pre-selected set of X-forms. If so, it is possible to learn the X-form $X$ successfully, so the processing associated with $X$. Thus, the success of a deep learning model depends on its set up. Having a good set up, the model will work well. Otherwise, no matter what data and what efforts, the model will not work well.  

Of course, besides the set up of model, learning methods are crucial. It is not easy to choose the right $X$ from  $\mathfrak{X}$ at all. We would like to point out the methods currently used indeed have some advantages:  
\begin{enumerate} [topsep=0pt,itemsep=-1ex,partopsep=1ex,parsep=1ex]
\item Methods act on Euclidean space, which is most easy to calculate, with many sophisticated algorithms, library, packages, and hardwares available. 
\item Methods are mostly doing linear algebraic calculus, which are easy to be parallelized. And, high parallelization is the key of its success. However, this advantage is build on this fact: no dynamic adaption. If dynamic adaption is used (such as recently introduced Capsule), this advantage might be lost.
\end{enumerate}

 {\bf Data for deep learning: }\\
As discussed above, the logical statement (X-form) is the core of deep learning. Without supporting data, a deep learning model could not reach a sophisticated logical statement (X-form). We defined data sufficiency in \cite{paper2}, which tells what kind of data can support one X-form (logical statement).  

Of course, the data sufficiency we defined is only the first step to understand data. Since deep learning approaching the desired X-form by some learning methods, it is easy to see that we need more data than just sufficient to support one X-form. The relationships here could be quite complicated, which would be the topic for further research. 

However, we can tell that data sufficient to support and sufficient to bound the desired X-form is the necessary condition for deep learning. In this sense, for so called big data for deep learning, we understand the necessity and lower bound.

{\bf Disadvantages of deep learning: }\\
Deep learning has some fundamental disadvantages from its root. We list some of them below: 
\begin{enumerate} [topsep=0pt,itemsep=-1ex,partopsep=1ex,parsep=1ex]
\item It is acting on huge parameter space, but, actual learning dynamics is on a fixed set of regions (which is equivalent to a set of X-forms). This indirectness makes every aspects of learning harder, especially, it is near impossible to know what is exactly happening in learning dynamics.
\item Successful learning needs data sufficient to support and to bound. This is very costly.
\item The structure of learning is setup by human. Once setup, structure (how many layers, how big a layer is, how layers fitting together, how to do convolution, etc) is not be able to change. This means that learning is restricted on a fixed group of regions, equivalent a fixed group of X-forms. If best X-form is not in this set, deep learning has no way to reach best X-form, no matter how big data are and how hard to try. Consequently, it is not universal learning machine.
\item It is very costly to embed X-forms into a huge parameter space. Perhaps, among all computing spend on learning, only a very small fraction is used on critical part, i.e. moving X-form to another, and most are just wasted.
\item Since there is no clear and reachable internal representation (due to the embedding), it will be very hard to do advanced learning, such as to unite all 5 learning methods together (see \cite{pedro}, \cite{paper2}).
\item Since there is no clear internal representation space, it is hard to define initial X-form, which is very essential to efficiency improving and several stages of learning.
\end{enumerate}


{\bf Looking forward to universal learning machine: }\\
Since deep learning model is not universal learning machine, naturally, we looking forward to universal learning machine. We discussed this in \cite{paper1} and \cite{paper2}. There, we proved that with certain capabilities, we can make universal learning machine. Also, we actually invented a concrete universal learning machine which is in the patent application process. We think universal learning machine has many advantages over deep learning model. There are many research works needed to be done for universal learning machine.


\begin{thebibliography}{99}
\bibitem{paper1} Chuy Xiong. Discussion on Mechanical Learning and Learning Machine, arxiv.org, 2016. \\ \htmladdnormallink{http://arxiv.org/pdf/1602.00198.pdf}{http://arxiv.org/pdf/1602.00198.pdf}

\bibitem{paper2} Chuy Xiong.  Descriptions of Objectives and Processes of Mechanical Learning, arxiv.org, 2017. \\ \htmladdnormallink{http://arxiv.org/abs/1706.00066.pdf}{http://arxiv.org/pdf/1706.00066.pdf}

\bibitem{pedro}  Pedro Domingos. The Master Algorithm, Talks at Google. \\ \htmladdnormallink{https://plus.google.com/117039636053462680924/posts/RxnFUqbbFRc}{https://plus.google.com/117039636053462680924/posts/RxnFUqbbFRc}

\bibitem{hinton} E. G. Hinton. Learning multiple layers of representation, Trends in Cognitive Sciences, Vol. 11, pp 428-434.. \htmladdnormallink{http://www.cs.toronto.edu/~hinton/absps/tics.pdf}{http://www.cs.toronto.edu/~hinton/absps/tics.pdf}

\bibitem{why} Henry W. Lin, Max Tegmark, David Rolnick. Why does deep and cheap learning work so well?, arxiv.org, 2016. \\ \htmladdnormallink{http://arxiv.org/pdf/1608.08225.pdf}{http://arxiv.org/pdf/1608.08225.pdf}


\bibitem{renormal} Pankaj Mehta, David J. Schwab, David Rolnick. An exact mapping between the Variational Renormalization Group and Deep Learning, arxiv.org, 2014. \\ \htmladdnormallink{http://arxiv.org/pdf/1410.3831.pdf}{http://arxiv.org/pdf/1410.3831.pdf}


\end{thebibliography}
\end{document}